\documentclass{article}

\PassOptionsToPackage{numbers, compress}{natbib}

\usepackage[final]{tackling_climate_workshop_style}

\usepackage[utf8]{inputenc} 
\usepackage[T1]{fontenc}    
\usepackage{hyperref}       
\usepackage{url}            
\usepackage{booktabs}       
\usepackage{amsfonts}       
\usepackage{nicefrac}       
\usepackage{microtype}      
\usepackage{amsmath}
\usepackage{graphicx}
\usepackage{wrapfig}
\usepackage{caption}
\usepackage{subcaption}

\title{Monitoring Sustainable Global Development Along Shared Socioeconomic Pathways}

\author{
    Michelle W.L. Wan \\
    University of Bristol, UK \\
    \texttt{mwlw3@cam.ac.uk}
    \And
    Jeffrey N. Clark \\
    University of Bristol, UK \\
    \texttt{jeff.clark@bristol.ac.uk} \\
    \And
    Edward A. Small \\
    University of Bristol, UK \\
    \texttt{edward.small@student.rmit.edu.au} \\
    \And
    Elena Fillola Mayoral \\
    University of Bristol, UK \\
    \texttt{elena.fillolamayoral@bristol.ac.uk} \\    
    \And
    Raúl Santos-Rodríguez \\
    University of Bristol, UK \\
    \texttt{enrsr@bristol.ac.uk}
}

\begin{document}

\maketitle

\begin{abstract}
Sustainable global development is one of the most prevalent challenges facing the world today, hinging on the equilibrium between socioeconomic growth and environmental sustainability. We propose approaches to monitor and quantify sustainable development along the Shared Socioeconomic Pathways (SSPs), including mathematically derived scoring algorithms, and machine learning methods. These integrate socioeconomic and environmental datasets, to produce an interpretable metric for SSP alignment. An initial study demonstrates promising results, laying the groundwork for the application of different methods to the monitoring of sustainable global development.
\end{abstract}

\section{Introduction}
To address the ongoing climate crisis, global socioeconomic progress must be balanced with environmental sustainability. However, assessing a region's overall development trajectory with this balance in mind poses a complex challenge. Progress is typically monitored with respect to warming limits of 1.5 $^\circ$C and 2 $^\circ$C using self-reported nationally determined contributions (NDCs) of emissions \cite{IPCCAR6WG3lecocq2022mitigation}. A study of European cities integrated environmental and socioeconomic datasets and applied machine learning techniques to assess emissions reduction efforts \cite{hsu2022predicting}. Despite substantial literature on country-level mitigation pathways, not all countries are well represented \cite{IPCCAR6WG3lecocq2022mitigation}. In climate change mitigation research, reference scenarios are used to evaluate possible low-carbon strategies; policymakers and researchers tend to focus on an individual scenario, which can limit the scope of understanding \cite{Grant2020}.

In 2017, the five Shared Socioeconomic Pathways (SSPs) were introduced: (1) Sustainability, (2) Middle of the Road, (3) Regional Rivalry, (4) Inequality, (5) Fossil-fueled Development \cite{Riahi2017_oecd_ssps, ONEILL2017169_SSPs}. These characterise the ways in which socioeconomic factors and atmospheric emissions may change in the coming century; analysis in 2020 identified over 1370 studies utilising the SSPs \cite{oneill2020achievements}. SSP projection datasets for complex, and sometimes competing, socioeconomic and environmental features are available from the year 2015, projected ahead to 2100. While the SSPs are presented as \textit{potential} pathways, rather than ideal pathways or targets, they can be used to contextualise current development trajectories. With this in mind, the SSP projections can be compared to observational data to evaluate the development sustainability of different regions relative to the established SSPs. With an increasing range of stakeholders using climate scenario information, it becomes ever more crucial to improve the communication of climate scenario research \cite{oneill2020achievements}. Thus, we propose the application of data science and machine learning methods to distill these existing datasets into a single, interpretable score. These methods will act as pivotal tools to assist decision-makers in addressing the climate crisis, and monitoring sustainable socioeconomic development.

\section{Proposed methods}
Here we propose the distillation of environmental and socioeconomical features into a single interpretable metric, to evaluate development against established scenarios. We propose initial investigation into mathematical methods: norm induced measures, to simply quantify the difference between observations and SSPs, and TraCE (Trajectory Counterfactual Explanation) scores \cite{clark2023trace}, which treat SSPs as counterfactual instances; as well as exploration of machine learning methods to better capture complex relationships. Additionally, scores from independent methods can be ensembled, to more robustly quantify SSP alignment than any single approach. These methods can be applied at a range of spatial scales, from regional to global.

\subsection{Norm induced measures}
Given a norm such as the Euclidean $\lVert f \rVert_2 = \sqrt{\int_\Omega f(x)^2 dx}$, we consider the difference between ground truth $u_i$ and target SSP $v_i$ for each feature $i$ as the error $e_i=u_i-v_i$. The norm measures the average error between truth and target by finding $\lVert e_i\rVert$ over the domain of interest $\Omega_i$ and dividing by time frame length (or discrete number of points). This measure is summed over all features (with weight $w_i$ for each feature), and divided by the sum of the weights to obtain a score:
\begin{equation}
    S = \frac{1}{\lVert w \rVert_2}\sum_{i=1}^n w_i\lVert  e_i \rVert_2
\end{equation}
This method allows for varying resolutions between features without interpolation. Weights can be selected to allow greater contribution of important measures to the score than other features. Using this method, if $S=0\implies e=0$ and therefore $u=v$, an SSP is followed perfectly. $S$ is therefore unbounded but strictly positive, with a larger value of $S$ indicating a larger value for $e$ and therefore worse alignment with an SSP.

Preliminary results are presented in Section \ref{sec:prelim_results}, with all features equally weighted.

\subsection{TraCE scores}
TraCE \cite{clark2023trace} outputs a score $S$ between $-1$ and $1$, derived from the combination of an angle metric $R_1(x_t, x^\prime_t)$ and a distance metric $R_2(x_t, x^\prime_t)$, weighted by $\lambda\in[0,1]$:
\begin{equation}
    S(x_t, x^\prime_t) = \lambda R_1(x_t, x^\prime_t) + (1 - \lambda) R_2(x_t, x^\prime_t)
\end{equation}
Positive values of $S$ indicate that a factual instance, $x_t$, representing historical observations, is moving towards a target, $x_{t}^\prime$, representing an SSP projection; negative values indicate movement away from the SSP.
Target point selection and $\lambda$ are user-adjustable parameters to be informed by domain expertise.
$\lambda$ can also be implemented as a learnable function from machine learning methods, or calibrated using the outputs of other scoring methods.

Informed by user priorities or a specific research question, we propose exploring input feature weighting, which can be assigned individually or by grouping features, for example into economic and environmental categories. An exploration of TraCE score presentation will identify the most useful summary statistics and visualisations for communicating results with stakeholders. 

Section \ref{sec:prelim_results} presents a preliminary study, with $\lambda = 0.9$, and all features weighted equally.

\subsection{Time series classification}
Using supervised machine learning methods, time series classification algorithms can be applied to labelled SSP projection datasets. Observational time series data for a given region can then be classified against the different SSP labels, with classification probability enabling a quantified comparison of a region's alignment with each SSP in turn. Approaches of interest include deep learning architectures, such as Long Short Term Memory (LSTM) networks \cite{karim2018lstm, karim2019multivariate}, and transformers \cite{zerveas2021transformer, jiang2022multi}.

\section{Preliminary study} \label{sec:prelim_results}

\begin{figure*}[t]
     \centering
        \begin{subfigure}[t]{0.49\textwidth}
         \centering
         \includegraphics[trim={1pt 1pt 1pt 1pt},clip,width=\textwidth]{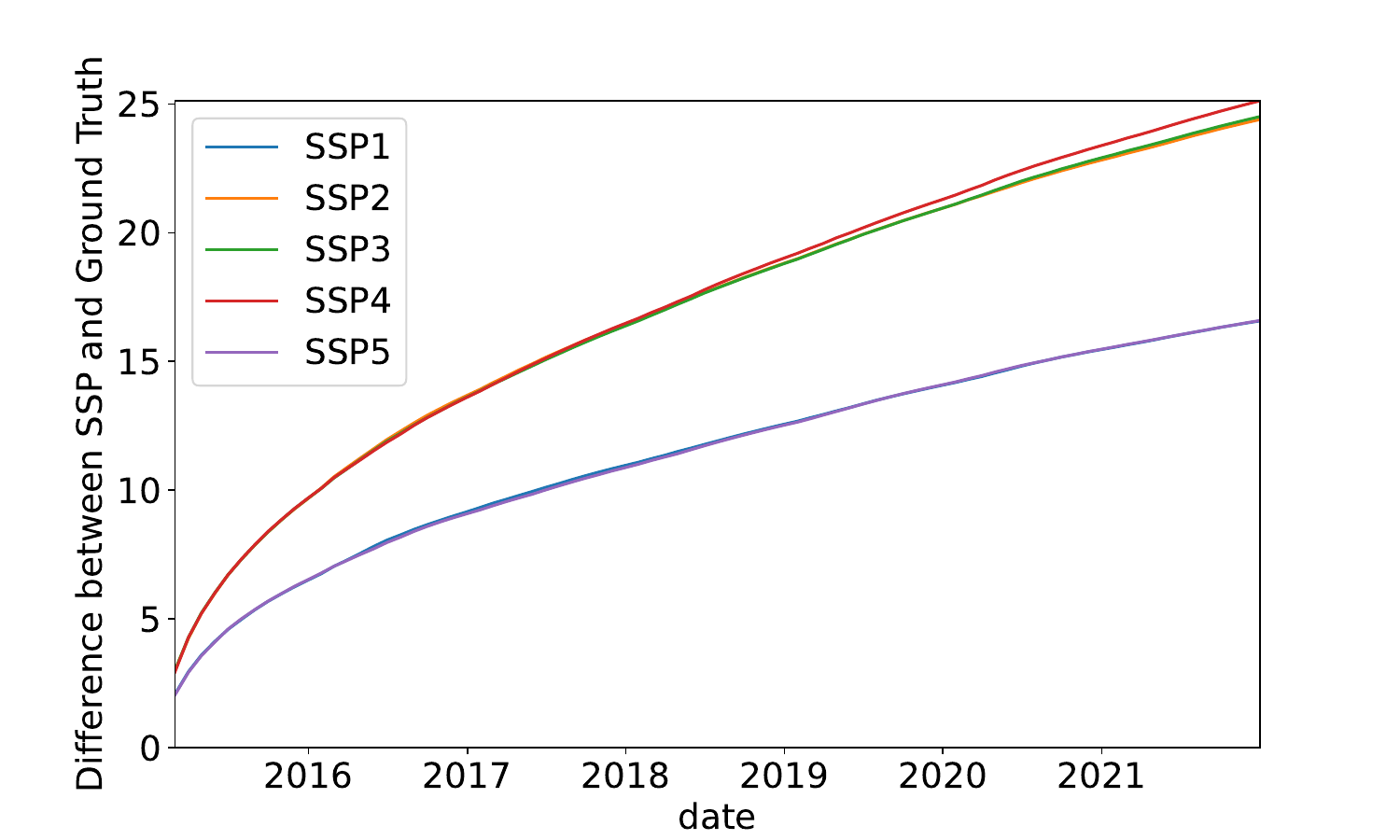}
          \caption{Norm induced measure over time for Brazil, for the period 2015-2022. Lower scores indicate closer alignment with a Shared Socioeconomic Pathway (SSP).}
          \label{fig:norm_plot}
     \end{subfigure}
     \hfill
     \begin{subfigure}[t]{0.49\textwidth}
         \centering
         \includegraphics[trim={1pt 1pt 1pt 1pt},clip,width=\textwidth]{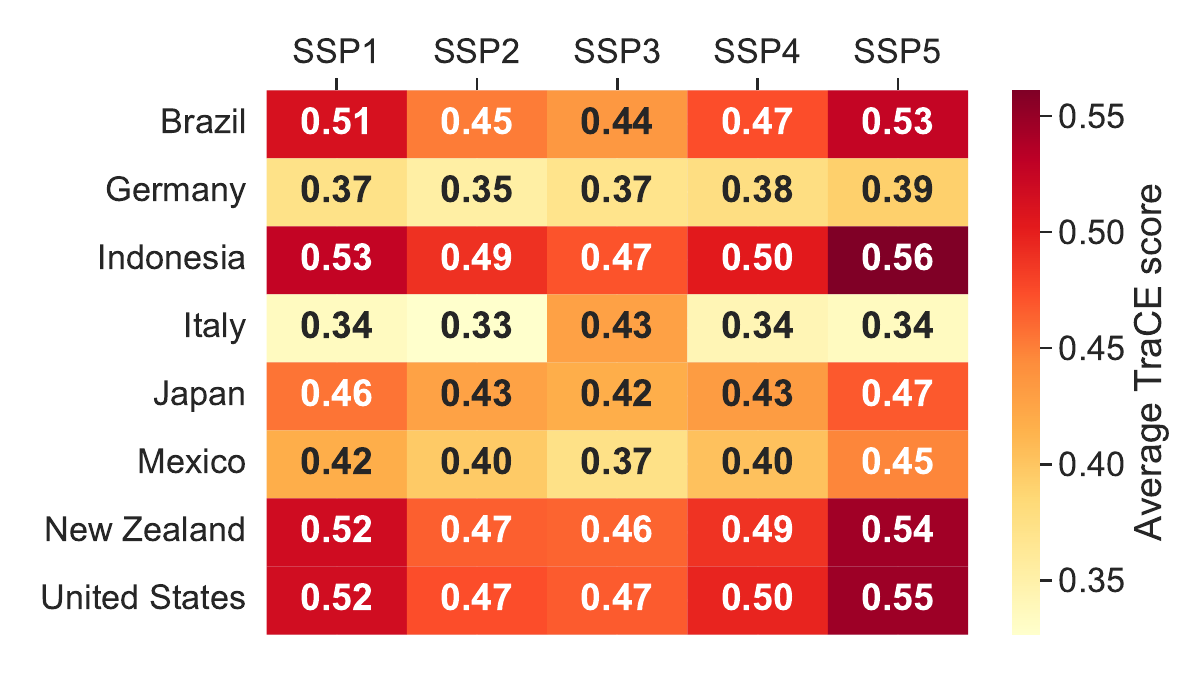}
         \caption{Average TraCE scores for eight countries, for the period 2015-2022. Higher scores indicate closer alignment with a Shared Socioeconomic Pathway (SSP).} 
         \label{fig:ssp_trace_heatmap}
     \end{subfigure}
     \caption{Preliminary results for two of the proposed methods}
\end{figure*}

An initial exploratory study leverages global data from 2015 to 2022. Five features from historical environmental (temperature, precipitation, methane) and socioeconomic (population, GDP) data are compared against corresponding SSP projections at each time point, to quantify country-level alignment with each SSP. By distilling alignment across these features into a single value, sustainable development trajectories can be quantified both within and between countries. Scores for a single regional time series are shown in Figure \ref{fig:norm_plot}, as the overall norm induced measures for Brazil. With this method, alignment for Brazil is strongest with SSP5 (Fossil-fueled Development) and SSP1 (Sustainability). TraCE scores are shown in Figure \ref{fig:ssp_trace_heatmap} for multiple regions, with countries generally aligning most strongly with SSP5. Germany scores lower across all SSPs, with little variation between different pathways, while Italy instead aligns most highly with SSP3 (Regional Rivalry).
Future work could investigate how alignment patterns may vary at different spatial scales, and how these may be explained by computing scores at the feature level.

 


\section{Responsible implementation and impact}
Here we propose the development of a framework to monitor sustainable development. With an interpretable approach to reconcile numerous and sometimes conflicting features, non-experts are able to quickly assess alignment with SSP scenarios. This is particularly useful because manual assessment of raw data becomes increasingly challenging as more features are included. These methods can therefore enable improved communication between stakeholder groups.

To ensure fully informed implementation, this work requires active engagement with experts across domains encompassing environmental, social, and economic sciences. Several factors will affect results, including the selection of data features and their weighting, and the model sources for SSP projection data. This in turn will affect the conclusions drawn about the trajectories of regions in alignment with the different SSPs. Reported results must therefore clearly discuss these considerations and the implications of choices made. Implementation must also explicitly acknowledge that these analyses do not attribute sole responsibility to specific regions for the SSP alignment outcomes. This is because the observed features used for monitoring one region's alignment can be influenced by the actions of other regions.

We envision this work as a component of a broader toolkit for applications such as monitoring real-time trajectories, and emissions simulation experiments, providing an output metric to quantify SSP alignment. This will enable decision-makers to effectively monitor current development trajectories, and evaluate the impact of possible actions in the context of the climate crisis.

\begin{ack}
MWLW, JNC, and RSR are funded by the UKRI Turing AI Fellowship [grant number EP/V024817/1]. EAS is funded by the ARC Centre of Excellence for Automated Decision-Making and Society (project number CE200100005), funded by the Australian Government through the Australian Research Council. Part of this work was done within the University of Bristol’s Machine Learning and Computer Vision (MaVi) Summer Research Program 2023. EFM is funded by a Google PhD Fellowship.
\end{ack}



\bibliographystyle{abbrvnat}
\bibliography{bib}

\end{document}